\documentclass[authoryear,a4paper,fleqn]{cas-sc}

\usepackage[authoryear]{natbib}

\usepackage{amssymb}

\usepackage[caption=false,font=footnotesize]{subfig}

\newcommand{\bbbr}{\mathbb{R}}
\newcommand{\bbbn}{{{{\mathbb{N}}}}}
\newcommand{\thickhline}{\noalign{\hrule height 0.8pt}}

\begin{document}
\let\WriteBookmarks\relax
\def\floatpagepagefraction{1}
\def\textpagefraction{.001}

\shorttitle{Reliable Prediction Intervals with Regression Neural Networks}

\shortauthors{Papadopoulos and Haralambous}

\title [mode = title]{Reliable Prediction Intervals with Regression Neural Networks}

\author{Harris Papadopoulos}[orcid=0000-0002-6839-6940]\cormark[1]
\ead{h.papadopoulos@frederick.ac.cy}
\cortext[1]{Corresponding author}

\author{Haris Haralambous}

\address{Computer Science and Engineering Department, Frederick University,\\7 Y. Frederickou St., Palouriotisa, Nicosia 1036, Cyprus.}

\begin{abstract}
This paper proposes an extension to conventional regression Neural Networks (NNs) 
for replacing the point predictions they produce with prediction intervals that 
satisfy a required level of confidence. Our approach follows a novel 
machine learning framework, called Conformal Prediction (CP), for assigning reliable 
confidence measures to predictions without assuming anything more than that the 
data are independent and identically distributed (i.i.d.). We evaluate the proposed method 
on four benchmark datasets and on the problem of predicting Total Electron Content (TEC),
which is an important parameter in trans-ionospheric links; for the latter we use
a dataset of more than 60000 TEC measurements collected over a period of 
11 years. Our experimental results show 
that the prediction intervals produced by our method are both well-calibrated 
and tight enough to be useful in practice.
\end{abstract}


\maketitle

\section{Introduction}\label{sec:intro}

Conformal Prediction (CP) is a novel framework for complementing the 
predictions of traditional machine learning algorithms with valid 
measures of their confidence. Confidence measures indicate the likelihood
of each prediction being correct and therefore provide the ability of 
making much more informed decisions. This makes them a highly desirable 
feature of the techniques developed for many real-world applications.

In this paper we develop a regression CP based on Neural Networks (NNs), which
is one of the most popular machine learning techniques. Some indicative 
fields in which NNs have been used with success are medicine, image processing, 
environmental modelling, robotics and the industry~\citep[see e.g.][]{mantzaris:anonsymb,iliadis:water,shuy:radar,iliadis:woodinfsci}.
In order to apply CP to NNs we follow a modified version of the 
original CP approach, called Inductive Conformal Prediction (ICP).
ICP was proposed by~\citet{papa:icm_rr} and~\citet{papa:icm_pr} 
in an effort to overcome
the computational inefficiency problem of CP. As demonstrated 
in the work of~\citet{papa:icpnn}, which describes ICP and its application 
to classification NNs, this computational inefficiency problem renders the 
original CP approach highly unsuitable for being coupled with NNs; 
and in general any method that requires long training times.

In the case of regression, instead of the point predictions produced by
conventional techniques, CPs produce prediction intervals that satisfy 
a given level of confidence. The important property of these 
intervals is that they are well-calibrated, meaning that in the long
run the intervals produced for some confidence level $1-\delta$ will
not contain the true label of an example with a relative frequency of 
at most $\delta$. Moreover, this is achieved without assuming anything
more than that the data are independent and identically 
distributed (i.i.d.), which is the typical assumption
of most machine learning methods. 

We first evaluate the proposed method on four benchmark datasets from 
the UCI~\citep{data:uci} and DELVE~\citep{data:delve} machine learning 
repositories. Then we apply it to the problem of predicting Total Electron
Content (TEC), which is an important parameter that represents a
quantitative measure of the detrimental effect of the ionosphere (an
ionised region in the upper atmosphere) on electromagnetic signals
from space-based systems propagating through it. Prediction of TEC 
enables mitigation techniques to be applied in order to reduce these undesirable
ionospheric effects on radar, communication, surveillance and navigation signals. 
For this reason, the use of NNs for TEC prediction was addressed in many
studies such as \citep{cander:tecnnet,maruyama:tecnnet,harala:tecnnet}. 
In this work we make one step further and provide prediction 
intervals, which make mitigation techniques more effective as they allow 
taking into account the highest possible TEC value at a desired confidence level.

The rest of the paper starts with an overview of related work on CP,
on the alternative ways of obtaining confidence information and
on the prediction of TEC and other Space Weather parameters in 
Section~\ref{sec:relwork}. This is followed by a brief description of 
the general idea behind CP and ICP in Section~\ref{sec:CP}, while 
Section~\ref{sec:nnricp} details the Nearest Neighbours Regression ICP
algorithm and gives the definition of a new normalized nonconformity measure.
In Section~\ref{sec:exp} the proposed method is evaluated experimentally on
four benchmark datasets. Subsequently, Section~\ref{sec:tec}, first describes 
the characteristics of TEC and the measurement data used in this study, while
it then presents its experimental results. Finally, Section~\ref{sec:conc}, 
gives the conclusions and future directions of this work.

\section{Related Work}\label{sec:relwork}

This section gives a synopsis of the work carried out on Conformal Prediction 
since it was first proposed, of the alternative ways that can be used for producing 
confidence information and their important drawbacks, and of the use of machine 
learning techniques for the prediction of parameters in ionospheric and generally 
Space Weather research.

\subsection{Conformal Prediction}

CP was initially proposed by \citet{gam:lbt} and later greatly 
improved by \citet{saunders:twcc}. In these papers 
CP was applied to Support Vector Machines for classification. Soon it 
started being applied to other popular classification algorithms, such as 
\emph{k}-Nearest Neighbours~\citep{proedrou:tcm_pr}, Decision Trees and 
Evolutionary Algorithms~\citep{lambrou:gacp}.
In the case of regression, where its application becomes 
more complicated, an initial attempt to apply it to Ridge Regression
was made by~\citet{melluish:tcm-rr}, while soon after a much better version was proposed 
by~\citet{nouret:tcm-rr}. Slightly later it was also applied to 
\emph{k}-Nearest Neighbours for Regression by~\citet{papa:nnr,papa:jairnnr}.

At the same time, work was being carried out on overcoming the 
computational inefficiency problem of CP, which was due to 
its transductive nature. After trying out some ways of improving
the efficiency of the transductive CP, such as ``competitive transduction''
and ``transduction with hashing'', a much more radical 
modification was made by moving to inductive inference. 
This modified version of CP, called Inductive Conformal 
Prediction (ICP) was proposed by~\citet{papa:icm_rr} 
for regression and by~\citet{papa:icm_pr} for classification.
ICP has also been applied to Neural Networks for classification 
in the work of~\citet{papa:icpnn}, where a computational complexity analysis
showing that ICPs are as efficient as their underlying algorithms
can be found.

To date CPs have been applied to a variety
of problems, such as the early detection of ovarian cancer~\citep{gam:preprot}, 
the classification of leukaemia subtypes~\citep{belloti:qualified},
the recognition of hypoxia electroencephalograms (EEGs)~\citep{zhang:hypoxia},
the prediction of plant promoters~\citep{gam:plantprom},
the diagnosis of acute abdominal pain~\citep{papa:eisaap},
the assessment of stroke risk~\citep{lambrou:stroke} and the 
estimation of effort for software projects~\citep{papa:soft:short}.

\subsection{Alternative Ways of Producing Confidence Information}

There are two other machine learning areas that can be used for
producing some kind of confidence information; these are the Bayesian framework 
and the theory of Probably Approximately Correct learning (PAC theory, \citealp{valiant:pac}).

The Bayesian framework can be used for producing methods that 
complement individual predictions with probabilistic measures of their quality.
These measures though, are based on some a priori assumptions about the 
distribution generating the data. If the correct prior is known, the measures 
produced by Bayesian methods are optimal. The problem is that for  
real world data the required knowledge is typically not available and as 
a result one is forced to assume the existence of some arbitrarily chosen
prior. In this case, since the assumed prior is most probably violated, the
outputs of Bayesian methods may become quite misleading. For example 
the prediction intervals output for the 95\% confidence level may contain the 
true label in much less than 95\% of the cases. This signifies a major failure, 
as we would expect confidence levels to bound the percentage of expected errors. 
\citet{papa:jairnnr} demonstrate experimentally this negative aspect of Bayesian 
techniques by applying Gaussian Process Regression~\citep{rasmussen:gp} to 
three benchmark datasets. A more detailed experimental comparison of Bayesian 
techniques and CP, that resulted in the same conclusion, for both classification 
and regression was performed by~\citet{melluish:bayes}.

On the other hand, PAC theory can be applied to an algorithm in order to produce 
upper bounds on the probability of its error with respect to some confidence level. 
Contrary to Bayesian techniques, PAC theory only assumes that the data are 
generated independently by some completely unknown distribution. In order for the 
bounds produced by PAC theory to be interesting in practice however, the dataset in 
question should be particularly clean. As this is rarely the case, the resulting 
bounds are typically very loose and therefore not very useful in practice. 
A demonstration of the crudeness of PAC bounds was performed by~\cite{nouretdinov:iid}. 
Furthermore, PAC theory has two additional drawbacks: 
(a) the majority of relevant results either involve large explicit constants or do 
not specify the relevant constants at all; (b) the bounds obtained by PAC theory 
are for the overall error and not for individual predictions.

All above problems are overcome by CPs, which in contrast to Bayesian techniques 
produce well-calibrated outputs as they are only based on the general i.i.d.\ assumption. 
Moreover, unlike PAC theory, they produce confidence measures that are useful in 
practice and are associated with individual predictions. Both the robustness of 
the resulting prediction intervals and their usefulness are demonstrated experimentally 
for the proposed methods in Sections~\ref{sec:exp} and~\ref{sec:exp2}.

\subsection{Space Weather Parameter Prediction}

Machine learning techniques have been applied widely in the last $18$ years for the 
specification, long-term prediction and short-term forecasting of Space Weather parameters 
and particularly ionospheric related parameters. The non-linear nature of a vast 
array of phenomena affecting the state and variability of Space Weather and the upper 
atmosphere within the whole chain of Solar Terrestrial effects represents a 
challenging field for the application of such techniques~\citep{lunds:nnets}.

Starting from the principal driver of Space Weather, solar activity, neural networks 
have been used to provide a functional representation of its benign 
state~\citep{macpherson:nnets, gleisner:disturbances}. In 
addition to neural networks, Support Vector Machines~\citep{gavrish:svm}, Time-Delay Neural 
Networks~\citep{gleisner:storms} and Elman Recurrent Neural Networks~\citep{wu:storms} 
have been applied to predict 
geomagnetic disturbances which are essentially short time scale consequence effects 
specific to solar activity transient phenomena. The nature and general morphology of 
these phenomena is very complex to model from first principles due to the numerous 
interactions involving various atmospheric layers.  

A number of studies have also been conducted concentrating on the prediction of TEC and 
other related ionospheric parameters important for telecommunication applications. 
These studies have dealt with short-term 
forecasting~\citep{cander:storm,liu:preliminary,koutroumbas:timeseries,strangeways:nearearth,stamper:nowcasting} 
and long-term prediction~\citep{maruyama:tecnnet,harala:tecnnet,agapitos:evoltec}
of such parameters as well as coping with missing data points~\citep{francis:missing}. 
Furthermore, it is important to note 
that work in this area is not limited to temporal variations of 
parameters, it also includes spatial ones~\citep{oyeyemi:globalfof2}.

\section{The Conformal Prediction Framework}\label{sec:CP}

This section gives a brief description of the CP
framework and its inductive version which is followed in this paper, for
more details the interested reader is referred to the book by~\citet{vovk:alrw}.
We are interested in making a prediction for the label of an example $x_{l+g}$, 
based on a set of training examples $\{(x_1, y_1), \dots, (x_l, y_l)\}$, 
where each $x_i\in \bbbr^d$  is the vector of attributes for example 
$i$ and $y_i\in \bbbr$ is the label of that example. Our only assumption 
is that all $(x_i,y_i)$, $i=1,2, \dots,$ have been generated independently 
from the same probability distribution (i.i.d.).

The idea behind CP is to assume every possible label $\tilde y$ 
of the example $x_{l+g}$ and check how likely it is that the extended set 
of examples
\begin{equation}
\label{eq:extset}
\{(x_1, y_1), \dots, (x_l, y_l), (x_{l+g}, \tilde y)\}
\end{equation}
is i.i.d. This
in effect will correspond to the likelihood of $\tilde y$ being the true 
label of the example $x_{l+g}$ since this is the only unknown value 
in~(\ref{eq:extset}). 

To do this we first assign a value $\alpha^{\tilde y}_i$ to each pair $(x_i, y_i)$
in~(\ref{eq:extset}) which indicates how strange, or nonconforming, this pair 
is for the rest of the examples in the same set. This value, called the 
\emph{nonconformity score} of the pair $(x_i,y_i)$, is calculated using a 
traditional machine learning algorithm, called the \emph{underlying algorithm} of 
the corresponding CP. More specifically, we train the underlying algorithm 
on~(\ref{eq:extset}) and generate the prediction rule
\begin{equation}
D_{\{(x_1, y_1), \dots, (x_l, y_l), (x_{l+g}, \tilde y)\}},
\end{equation}
which maps any input pattern $x_i$ to a predicted label $\hat y_i$. The 
nonconformity score of each pair $(x_i,y_i):y = 1,\dots,l,l+g$ is then measured 
as the degree of disagreement between the prediction
\begin{equation}
\label{eq:rulepred}
\hat y_i = D_{\{(x_1, y_1), \dots, (x_l, y_l), (x_{l+g}, \tilde y)\}} (x_i)
\end{equation}
and the actual label $y_i$; note that in the case of the pair $(x_{l+g}, \tilde y)$ 
the actual label is replaced by the assumed label $\tilde y$. The function used 
for measuring this degree of disagreement is called 
the \emph{nonconformity measure} of the CP. Notice that a change in the assumed 
label $\tilde y$ affects all predictions~(\ref{eq:rulepred}) since it is part 
of the underlying algorithm's training set.

The nonconformity score $\alpha^{\tilde y}_{l+g}$ is then compared to the nonconformity 
scores of all other examples to find out how unusual the pair $(x_{l+g}, \tilde y)$ is according 
to the nonconformity measure used. This comparison is performed with the function
\begin{equation}
\label{eq:pvalue}
  p(\{(x_1, y_1), \dots, (x_l, y_l), (x_{l+g}, \tilde y)\}) = \frac{\#\{i = 1, \dots, l, l+g : \alpha^{\tilde y}_i \geq \alpha^{\tilde y}_{l+g}\}}{l+1},
\end{equation} 
which calculates the portion of examples in~(\ref{eq:extset}) that are equally 
or more nonconforming than $(x_{l+g}, \tilde y)$. The output of this function, which lies 
between $\frac{1}{l+1}$ and 1, is called the p-value of $\tilde y$, also denoted 
as $p(\tilde y)$. An important property of (\ref{eq:pvalue}) 
is that $\forall \delta\in [0, 1]$ and for all probability distributions $P$ on $Z$,
\begin{equation}
\label{eq:validity}
  P\{\{(x_1,y_1), \dots, (x_l, y_l), (x_{l+g},y_{l+g})\}:p(y_{l+g}) \leq \delta\}\leq \delta,
\end{equation}
in other words for i.i.d.\ data, the probability of the p-value for the true label $y_{l+g}$ 
being less than or equal to any given threshold $\delta$ is less than or equal to $\delta$;
a proof can be found in the book by~\citet{vovk:alrw}. This makes it a valid test of randomness 
with respect to the i.i.d.\ model. According to this property, if $p(\tilde y)$ is under some 
very low threshold, say $0.05$, this means that $\tilde y$ is highly unlikely 
of being the true label as the probability of such an event is at most $5\%$ 
if~(\ref{eq:extset}) is i.i.d. 

Assuming it were possible to calculate the p-value of every possible label
following the above procedure, we could then exclude all labels with a p-value 
under some very low threshold, or \emph{significance level}, $\delta$ and have 
at most $\delta$ chance of being wrong. Consequently, given a confidence level
$1 - \delta$ a regression CP outputs the set
\begin{equation}
\label{eq:predregion}
	\{ \tilde y : p(\tilde y) > \delta \},
\end{equation}
in other words the interval containing all labels that have a p-value 
greater than $\delta$.
Of course it is impossible to explicitly consider every possible 
label $\tilde y \in \bbbr$, so regression CPs follow a different 
approach which makes it possible to compute the prediction interval (\ref{eq:predregion}). 
This approach is described by~\citet{nouret:tcm-rr} for Ridge Regression 
and by~\citet{papa:nnr} for $k$-Nearest Neighbours Regression.

\subsection{Inductive Conformal Prediction}\label{sec:ICP}

The only drawback of the original CP approach is that due to its transductive
nature all its computations, including training the underlying algorithm, 
have to be repeated for each assumed label of every new test example. This makes it 
very computationally inefficient especially for algorithms that require long 
training times such as NNs. Furthermore, in the case of regression where it is 
impossible to explicitly consider every possible label of the new example, the 
approach followed by~\citet{nouret:tcm-rr} and~\citet{papa:nnr} for computing 
the prediction interval (\ref{eq:predregion}) can only be employed
if it is possible to calculate how a change in $\tilde y$ will affect the 
predictions produced by~(\ref{eq:rulepred}) for all examples in~(\ref{eq:extset})
and consequently the resulting nonconformity scores 
$\alpha^{\tilde y}_1, \dots, \alpha^{\tilde y}_l, \alpha^{\tilde y}_{l+g}$.
ICP is based on the same theoretical foundations described above, but performs 
inductive rather than transductive inference. As a result ICP is almost as 
efficient as its underlying algorithm~\citep{papa:icpnn} and it can be combined 
with any conventional regression method.

ICP splits the training set (of size $l$) into two smaller sets, the 
\emph{proper training set} with $m<l$ examples and the \emph{calibration set} 
with $q:=l-m$ examples. It then uses the proper training set for training
its underlying algorithm and the calibration set for calculating the p-value 
of each possible label $\tilde y$. More specifically, it trains the underlying 
algorithm on $(x_1, y_1), \dots, (x_m, y_m)$ to generate the prediction rule 
\begin{equation}\label{ICPpredrule}
D_{\{(x_1, y_1), \dots, (x_m, y_m)\}},
\end{equation}
and calculates the nonconformity score $\alpha_{m+i}$ of each example in the 
calibration set $(x_{m+i}, y_{m+i}), i = 1, \dots, q$ as the degree of disagreement 
between the prediction
\begin{equation}
\label{eq:rulepred2}
\hat y_{m+i} = D_{\{(x_1, y_1), \dots, (x_m, y_m)\}} (x_{m+i})
\end{equation}
and the actual label $y_{m+i}$. This needs to be done only once as now $x_{l+g}$ 
is not included in the training set of the underlying algorithm. From this
point on, it only needs to compute the prediction 
\begin{equation}
\label{eq:rulepred3}
\hat y_{l+g} = D_{\{(x_1, y_1), \dots, (x_m, y_m)\}} (x_{l+g})
\end{equation}
for each new example $x_{l+g}$ and calculate the nonconformity score 
$a^{\tilde y}_{l+g}$ of the pair $(x_{l+g}, \tilde y)$ for every 
possible label $\tilde y$ as the degree of disagreement between itself the 
prediction $\hat y_{l+g}$. The p-value of $\tilde y$ can now be 
calculated as
\begin{equation}
\label{eq:pvalueicp}
  p(\tilde y) = \frac{\#\{i = m+1, \dots, m+q, l+g : \alpha_i \geq \alpha^{\tilde y}_{l+g}\}}{q+1}.
\end{equation} 
Again it is impossible to explicitly go through every possible label 
$\tilde y \in \bbbr$ to calculate its p-value, but it is possible to compute 
the prediction interval (\ref{eq:predregion}) as we show in the next section.

\section{Neural Networks Regression ICP}\label{sec:nnricp}

In order to use ICP in conjunction with some traditional algorithm we first have 
to define a nonconformity measure. Recall that a nonconformity measure is a 
function that measures the disagreement between the actual label $y_i$ and the 
prediction $\hat y_i$ produced by the prediction rule (\ref{ICPpredrule}) of the 
underlying algorithm for the example $x_i$. In the case of regression 
this can be easily defined as the absolute difference between the two
\begin{equation}
\label{eq:nm1}
  \alpha_i = |y_i - \hat y_i|.
\end{equation} 
We first describe the Neural Networks Regression ICP (NNR ICP) algorithm with this
measure and then define a \emph{normalized nonconformity measure}, which 
has the effect of producing tighter prediction intervals by taking into account 
the expected accuracy of the underlying NN on each example.

The first steps of the NNR ICP algorithm follow exactly 
the general scheme given in Section~\ref{sec:ICP}: 
\begin{itemize}
\item Split the training set $\{(x_1, y_1), \dots, (x_l, y_l)\}$ into two subsets:
      \begin{itemize}
      \item the proper training set: $\{(x_1, y_1), \dots, (x_m, y_m)\}$, and
      \item the calibration set: $\{(x_{m+1}, y_{m+1}), \dots, (x_{m+q}, y_{m+q})\}$.
      \end{itemize}
\item Use the proper training set to train the NN.
\item For each pair $(x_{m+i}, y_{m+i}), i=1, \dots, q$ in the calibration set:
      \begin{itemize}
      \item supply the input pattern $x_{m+i}$ to the trained NN to obtain the prediction $\hat y_{m+i}$ and
      \item calculate the nonconformity score $\alpha_{m+i}$ with (\ref{eq:nm1}).
      \end{itemize}
\end{itemize}
At this point however, it becomes impossible to follow the general ICP scheme as 
there is no way of trying out all possible labels $\tilde y \in \bbbr$ in order to
calculate their nonconformity score and p-value. Notice though that both the 
nonconformity scores of the calibration set examples $\alpha_{m+1}, \dots, \alpha_{m+q}$ 
and the prediction $\hat y_{l+g}$ of the trained NN for the new example
$x_{l+g}$ will remain fixed as we change the assumed label $\tilde y$. The only value 
that will change is the nonconformity score 
\begin{equation}
\alpha^{\tilde y}_{l+g} = |\tilde y - \hat y_{l+g}|.
\end{equation}
Thus $p(\tilde y)$ will only change at the points where $\alpha^{\tilde y}_{l+g} = \alpha_{m+i}$
for some $i = 1, \dots, q$. As a result, for a given confidence level $1 - \delta$ we 
only need to find the biggest $\alpha_{m+i}$ such that when $\alpha^{\tilde y}_{l+g} = \alpha_{m+i}$ 
then $p(\tilde y) > \delta$, which will give us the maximum and minimum $\tilde y$
that have a p-value greater than $\delta$ and consequently the beginning and end
of the corresponding prediction interval. More specifically, after calculating the 
nonconformity scores of all calibration examples the NNR~ICP algorithm 
continues as follows:
\begin{itemize}
\item Sort the nonconformity scores of the calibration examples in descending
order obtaining the sequence
 \begin{equation}\label{eq:sortedalphas}
          \alpha_{(m+1)}, \dots, \alpha_{(m+q)}.
 \end{equation}
\item For each new test example $x_{l+g}$:
      \begin{itemize}
      \item supply the input pattern $x_{l+g}$ to the trained NN to obtain the prediction $\hat y_{l+g}$ and
      \item output the prediction interval
      \begin{equation}\label{eq:ICPregions}
               (\hat y_{l+g} - \alpha_{(m+s)}, \hat y_{l+g} + \alpha_{(m+s)}),
      \end{equation}
      where
      \begin{equation}\label{eq:ICPregions-s}
               s = \lfloor \delta (q + 1) \rfloor.
      \end{equation}      
      \end{itemize}
\end{itemize}

An important parameter of the ICP algorithm is the number $q$ of training examples 
that will be allocated to the calibration set and the nonconformity scores of which 
will be used by the ICP to generate its prediction intervals. This number should only 
correspond to a small portion of the training set, as in the opposite case the removal 
of these examples will result in a significant reduction to the predictive ability of 
the underlying NN and consequently to wider than desirable prediction intervals. 
As we are mainly interested in the confidence levels of $99\%$ and $95\%$, the 
calibration sizes we use are of the form $q = 100n - 1$, 
where $n \in \bbbn$ (see~(\ref{eq:ICPregions-s})).

\subsection{A Normalized Nonconformity Measure}

We extend nonconformity measure definition (\ref{eq:nm1}) by normalizing it 
with the predicted accuracy of the underlying NN on the given example.
This leads to prediction intervals that are larger for the ``difficult'' 
examples and smaller for the ``easy'' ones. As a result the ICP can 
satisfy the required confidence level with intervals that are on average 
tighter. 

Our new measure is defined as
\begin{equation}
\label{eq:nm2}
	\alpha_i = \frac{|y_i - \hat y_i|}{\exp(\mu_i) + \beta},
\end{equation} 
where $\mu_i$ is the prediction of the value $\ln(|y_i - \hat y_i|)$ produced by 
a linear NN trained on the proper training patterns and $\exp$ is the exponential 
function. In effect after training the underlying NN of the ICP we calculate 
the residuals $|y_j - \hat y_j|$ for all proper training examples $j = 1, \dots, m$ 
and train a linear NN on the pairs $(x_j, \ln(|y_j - \hat y_j|))$ producing
the prediction rule
\begin{equation}\label{eq:linearNNrule}
D_{\{(x_1, \ln(|y_1 - \hat y_1|)), \dots, (x_m, \ln(|y_m - \hat y_m|))\}}.
\end{equation}
Then $\mu_i$ is the prediction 
\begin{equation}
\mu_i = D_{\{(x_1, \ln(|y_1 - \hat y_1|)), \dots, (x_m, \ln(|y_m - \hat y_m|))\}}(x_i).
\end{equation} 
of this linear NN for the input pattern $x_i$. The parameter $\beta \geq 0$ 
controls the sensitivity of the measure to changes of $\mu_i$, since the 
latter depends on the range of possible labels and the complexity 
of the problem in question.

We use a linear rather than a more complex NN as we want the prediction 
rule~(\ref{eq:linearNNrule}) to capture only the important variation of the 
loss of the underlying NN and not be affected by small changes, which are mainly 
due to noise. Besides linear NN are much faster to train, which means that the 
computational efficiency of the ICP is not affected by much.
We also use the logarithmic instead 
of the direct scale to ensure that the estimate is always positive.

When using (\ref{eq:nm2}) as nonconformity measure the prediction interval
produced by the ICP for each new pattern $x_{l+g}$ becomes
\begin{equation}\label{eq:ICPregions2}
	(\hat y_{l+g} - \alpha_{(m+s)}(\exp(\mu_{l+g}) + \beta), \hat y_{l+g} + \alpha_{(m+s)}(\exp(\mu_{l+g}) + \beta)),
\end{equation}
where again $s = \lfloor \delta (q + 1) \rfloor$.

\section{Experimental Evaluation on Benchmark Datasets}\label{sec:exp}

We tested the proposed method on four benchmark datasets from the UCI~\citep{data:uci} 
and DELVE~\citep{data:delve} repositories:
\begin{itemize}
\item \emph{Boston Housing}, which lists the median house prices for $506$ 
	different areas of Boston MA in \$1000s. Each area is described by 
	13 attributes such as pollution and crime rate.
\item \emph{Abalone}, which concerns the prediction of the age of abalone from physical
	measurements. The data set consists of 4177 examples described by 8 attributes 
	such as diameter, height and shell weight.
\item \emph{Computer Activity}, which is a collection of a computer 
	systems activity measures
	from a Sun SPARCstation 20/712 with 128 Mbytes of memory running in a multi-user 
	university department. It consists of 8192 examples of 12 measured values, such as
	the number of system buffer reads per second and the number of system call writes 
	per second, at random points in time. The task is to predict the portion of time 
	that the cpus run in user mode, ranging from 0 to 100. We used the \emph{small} 
	variant of the data set which contains only 12 of the 21 attributes.
\item \emph{Bank}, which was generated from simplistic simulator of the 
	queues in a series of banks. The task is to predict the rate of rejections, 
	i.e.\ the fraction of customers that are turned away from the bank because 
	all the open tellers have full queues. The data set consists of 8192 examples
	described by 8 attributes like area population size and maximum possible length 
	of queues. We used the \emph{8nm} variant of the data set which contains 8 of 
    the 32 attributes, and is highly non-linear with moderate noise.
\end{itemize}

Before conducting our experiments the attributes of all datasets were normalized 
to a minimum value of -1 and a maximum value of 1. Our experiments followed a fold 
cross-validation process; each dataset was split randomly into $k$ folds of almost 
equal size and our experiments were repeated $k$ times each time using one of the 
$k$ folds as test set and the other $k-1$ folds as training set. In order to ensure 
that the results reported here do not depend on the particular split of the dataset
into the $k$ folds or in the particular choice of calibration examples, this process 
was repeated 10 times with different permutations of the examples. Based on their 
sizes the Boston Housing and Abalone datasets were split into 10 and 4 folds respectively, 
while the other two were splint into 2 folds. The calibration set sizes were 
set to $q = 100n - 1$ (see Section~\ref{sec:nnricp}), where $n$ was chosen so that $q$
was approximately $1/10$th of each dataset's training size; in the case of the 
Boston Housing data set the smallest value $n=1$ was used due to its small size.

\begin{table}
  \small
  \centering
  \begin{tabular}{rcccc} \hline\noalign{\smallskip}
                    & Boston  &         & Computer &              \\
                    & Housing & Abalone & Activity & \,\,Bank\,\, \\ \noalign{\smallskip}\thickhline\noalign{\smallskip}
  Examples          & 506 & 4177 & 8192 & 8192 \\
  Attributes        & 13  & 8    & 12   & 8    \\
  Label range       & 45  & 28   & 99   & 0.48 \\
  Folds             & 10  & 4    & 2    & 2    \\
  Hidden Neurons    & 7   & 8    & 17   & 13   \\
  Calibration size  & 99  & 299  & 399  & 399  \\\noalign{\smallskip}\hline
  \end{tabular}
  \caption{Main characteristics and experimental setup for each data set.}
  \label{tab:datasets}
\end{table}

\begin{table}
  \small
  \centering
  \begin{tabular}{rrccccccc} \hline\noalign{\smallskip}
                    & & \multicolumn{3}{c}{Original NNR} & & \multicolumn{3}{c}{NNR ICP} \\ \noalign{\smallskip}
                    & & RMSE  & &  CC   & & RMSE  & &  CC   \\ \noalign{\smallskip}\thickhline\noalign{\smallskip}
  Boston Housing    & & 4.059 & & 0.900 & & 4.307 & & 0.885 \\
  Abalone           & & 2.091 & & 0.761 & & 2.099 & & 0.759 \\
  Computer Activity & & 3.105 & & 0.986 & & 3.249 & & 0.984 \\
  Bank              & & 0.019 & & 0.953 & & 0.019 & & 0.950 \\\noalign{\smallskip}\hline
  \end{tabular}
  \caption{Point prediction performance of NNR ICP and its underlying Neural Network.}
  \label{tab:rmse}
\end{table}

\begin{table}[t]\footnotesize
  \centering
  \begin{tabular}{|c|rrr|rrr|rrr|} \hline
             & \multicolumn{3}{|c|}{} 
             & \multicolumn{3}{|c|}{Interdecile} & \multicolumn{3}{|c|}{Percentage} \\
             \multicolumn{1}{|c|}{Nonconformity} & \multicolumn{3}{|c|}{Median Width} 
             & \multicolumn{3}{|c|}{Mean Width} & \multicolumn{3}{|c|}{of Errors (\%)} \\
             \multicolumn{1}{|c|}{Measure} & \multicolumn{1}{|c}{90\%} & \multicolumn{1}{c}{95\%}
             & \multicolumn{1}{c|}{99\%} 
             & \multicolumn{1}{|c}{90\%} & \multicolumn{1}{c}{95\%}
             & \multicolumn{1}{c|}{99\%}
             & \multicolumn{1}{|c}{90\%} & \multicolumn{1}{c}{95\%}
             & \multicolumn{1}{c|}{99\%} \\ \hline
             (\ref{eq:nm1})                 & 12.44 & 17.18 & 39.32 & 12.49 & 17.07 & 38.68 & 10.18 & 4.66 & 0.87 \\
             (\ref{eq:nm2}) - $\beta = 0$   & 11.93 & 16.82 & 35.81 & 12.48 & 17.67 & 39.91 & 10.26 & 5.02 & 1.09 \\
             (\ref{eq:nm2}) - $\beta = 0.5$ & 11.67 & 16.13 & 32.19 & 12.03 & 16.74 & 34.94 & 10.16 & 4.94 & 1.03 \\ \hline  
\end{tabular}
  \caption{Tightness and empirical reliability results for the Boston Housing Dataset.}
  \label{tab:bostonres}
\end{table}

\begin{table}[t]\footnotesize
  \centering
  \begin{tabular}{|c|rrr|rrr|rrr|} \hline
             & \multicolumn{3}{|c|}{} 
             & \multicolumn{3}{|c|}{Interdecile} & \multicolumn{3}{|c|}{Percentage} \\
             \multicolumn{1}{|c|}{Nonconformity} & \multicolumn{3}{|c|}{Median Width} 
             & \multicolumn{3}{|c|}{Mean Width} & \multicolumn{3}{|c|}{of Errors (\%)} \\
             \multicolumn{1}{|c|}{Measure} & \multicolumn{1}{|c}{90\%} & \multicolumn{1}{c}{95\%}
             & \multicolumn{1}{c|}{99\%} 
             & \multicolumn{1}{|c}{90\%} & \multicolumn{1}{c}{95\%}
             & \multicolumn{1}{c|}{99\%}
             & \multicolumn{1}{|c}{90\%} & \multicolumn{1}{c}{95\%}
             & \multicolumn{1}{c|}{99\%} \\ \hline
             (\ref{eq:nm1})                 & 6.60 & 9.08 & 14.97 & 6.64 & 9.07 & 14.89 & 10.01 & 5.02 & 0.91 \\
             (\ref{eq:nm2}) - $\beta = 0$   & 5.48 & 7.31 & 12.69 & 5.73 & 7.63 & 13.28 & 9.86 & 4.89 & 0.85 \\
             (\ref{eq:nm2}) - $\beta = 0.5$ & 5.65 & 7.41 & 12.47 & 5.80 & 7.59 & 12.78 & 10.02 & 5.00 & 0.88 \\ \hline  
\end{tabular}
  \caption{Tightness and empirical reliability results for the Abalone Dataset.}
  \label{tab:abaloneres}
\end{table}

\begin{table}[t]\footnotesize
  \centering
  \begin{tabular}{|c|rrr|rrr|rrr|} \hline
             & \multicolumn{3}{|c|}{} 
             & \multicolumn{3}{|c|}{Interdecile} & \multicolumn{3}{|c|}{Percentage} \\
             \multicolumn{1}{|c|}{Nonconformity} & \multicolumn{3}{|c|}{Median Width} 
             & \multicolumn{3}{|c|}{Mean Width} & \multicolumn{3}{|c|}{of Errors (\%)} \\
             \multicolumn{1}{|c|}{Measure} & \multicolumn{1}{|c}{90\%} & \multicolumn{1}{c}{95\%}
             & \multicolumn{1}{c|}{99\%} 
             & \multicolumn{1}{|c}{90\%} & \multicolumn{1}{c}{95\%}
             & \multicolumn{1}{c|}{99\%}
             & \multicolumn{1}{|c}{90\%} & \multicolumn{1}{c}{95\%}
             & \multicolumn{1}{c|}{99\%} \\ \hline
             (\ref{eq:nm1})                 & 9.75 & 12.34 & 19.86 & 9.76 & 12.46 & 20.35 & 10.02 & 5.32 & 1.01 \\
             (\ref{eq:nm2}) - $\beta = 0$   & 8.81 & 11.01 & 17.94 & 8.98 & 11.21 & 18.44 & 10.24 & 5.43 & 1.13 \\
             (\ref{eq:nm2}) - $\beta = 0.5$ & 8.78 & 10.97 & 17.28 & 8.90 & 11.12 & 17.59 & 10.18 & 5.35 & 1.00 \\ \hline  
\end{tabular}
  \caption{Tightness and empirical reliability results for the Computer Activity Dataset.}
  \label{tab:cpures}
\end{table}

\begin{table}[t]\footnotesize
  \centering
  \begin{tabular}{|c|rrr|rrr|rrr|} \hline
             & \multicolumn{3}{|c|}{} 
             & \multicolumn{3}{|c|}{Interdecile} & \multicolumn{3}{|c|}{Percentage} \\
             \multicolumn{1}{|c|}{Nonconformity} & \multicolumn{3}{|c|}{Median Width} 
             & \multicolumn{3}{|c|}{Mean Width} & \multicolumn{3}{|c|}{of Errors (\%)} \\
             \multicolumn{1}{|c|}{Measure} & \multicolumn{1}{|c}{90\%} & \multicolumn{1}{c}{95\%}
             & \multicolumn{1}{c|}{99\%} 
             & \multicolumn{1}{|c}{90\%} & \multicolumn{1}{c}{95\%}
             & \multicolumn{1}{c|}{99\%}
             & \multicolumn{1}{|c}{90\%} & \multicolumn{1}{c}{95\%}
             & \multicolumn{1}{c|}{99\%} \\ \hline
             (\ref{eq:nm1})                 & 0.059 & 0.084 & 0.137 & 0.058 & 0.083 & 0.144 & 10.21 & 4.86 & 1.07 \\
             (\ref{eq:nm2}) - $\beta = 0$   & 0.038 & 0.048 & 0.078 & 0.042 & 0.053 & 0.087 & 9.96 & 5.16 & 1.05 \\
             (\ref{eq:nm2}) - $\beta = 0.5$ & 0.058 & 0.083 & 0.136 & 0.058 & 0.083 & 0.139 & 10.18 & 4.83 & 1.06 \\ \hline  
\end{tabular}
  \caption{Tightness and empirical reliability results for the Bank Dataset.}
  \label{tab:bankres}
\end{table}
\normalsize

The underlying NN had a fully connected two-layer structure. The hidden layer 
consisted of neurons with hyperbolic tangent activation functions, while the
output layer consisted of a single neuron with a linear activation function. 
The number of hidden neurons was determined by trial and error by performing 
a fold cross-validation process with the
original NN predictor on 10 different random permutations than the ones used
for evaluating the ICP. The training algorithm used was the Levenberg-Marquardt 
backpropagation algorithm with early stopping based on a validation set created 
from 10\% of the proper training examples. In an effort to avoid local 
minima 10 NNs were trained with different random initialisations 
and the one that performed best on the validation set was selected 
for being applied to the calibration and test examples. 

The number of examples and attributes of each dataset and the width of its
range of labels, together with the number of folds $k$, calibration set size $q$ 
and number of hidden units used in our experiments are reported in Table~\ref{tab:datasets}.
In the case of nonconformity measure (\ref{eq:nm2}) we experimented with $\beta = 0$ and 
$\beta = 0.5$ for all datasets to explore the difference that the addition of 
this parameter makes. It is worth to note however, that somewhat tighter prediction
intervals can be obtained by adjusting $\beta$ for each dataset. We chose not
to do this here so as to show that the huge improvement in prediction interval 
widths resulting from the use of this nonconformity measure does not depend on 
fine tuning this parameter.

Table~\ref{tab:rmse} reports the performance of the point predictions of our method 
in terms of its Root Mean Squared Error (RMSE) and the Correlation Coefficient (CC) 
between the predicted and actual values and compares them to those of its 
underlying NN. This table basically shows the effect that the removal of the 
calibration examples has on the performance of the NN, since this is the only 
difference between the two as far as point predictions are concerned. The values 
presented in this table show that the performance decrease is not significant.
This is a small prise that we have to pay for the much more informative outputs 
of the ICP. 

Since the advantage of our method is that it produces prediction intervals,
the main aim of our experiments was to check their tightness, and therefore usefulness, 
and their empirical reliability. To this end the first two parts of 
Tables~\ref{tab:bostonres}-\ref{tab:bankres} report the median and interdecile mean 
widths of the prediction intervals produced for the four datasets when using 
nonconformity measures (\ref{eq:nm1}) and (\ref{eq:nm2}) with $\beta$ set 
to $0$ and $0.5$ for the $90\%$, $95\%$ and $99\%$ confidence levels. 
We chose to report the median and interdecile mean values 
instead of the mean for evaluating prediction interval tightness so as 
to avoid the strong impact of a few extremely large or extremely small 
intervals. The third and last part of Tables~\ref{tab:bostonres}-\ref{tab:bankres} 
reports the percentage of errors made each time, which is in fact the percentage
of intervals that did not contain the true label of the example.

\begin{figure}
	\centering
		\subfloat[Boston Housing]{\includegraphics[trim = 2mm 6mm 0mm 0mm, clip, width=6.5cm]{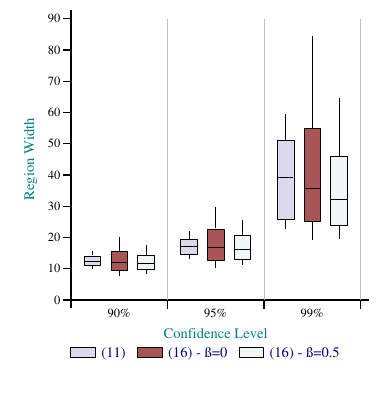}\label{fig:boston}}
		\subfloat[Abalone]{\includegraphics[trim = 2mm 6mm 0mm 0mm, clip, width=6.5cm]{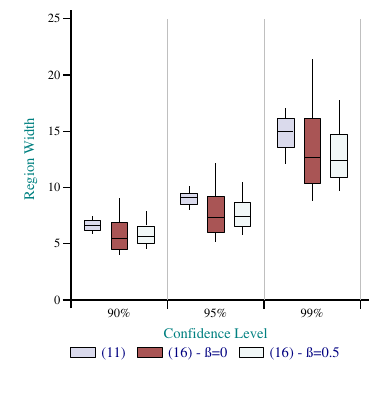}\label{fig:abalone}}\\
		\subfloat[Computer Activity]{\includegraphics[trim = 2mm 6mm 0mm 0mm, clip, width=6.5cm]{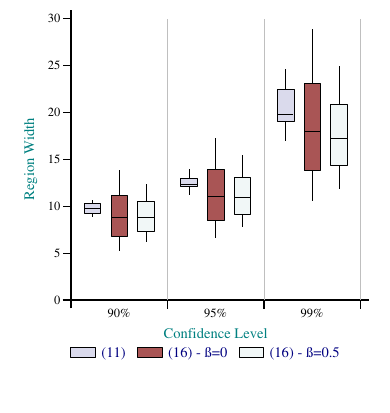}\label{fig:cpu}}
		\subfloat[Bank]{\includegraphics[trim = 2mm 6mm 0mm 0mm, clip, width=6.5cm]{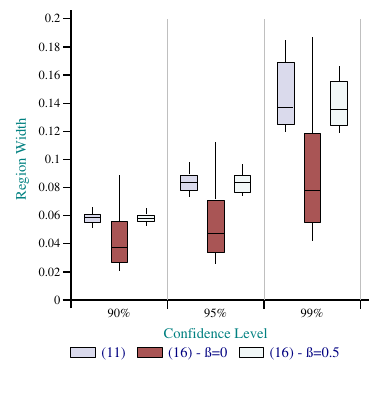}\label{fig:bank}}
\caption{Prediction interval width distribution for each dataset.}
\label{fig:widths}
\end{figure}

More information on the tightness of the obtained prediction intervals are given 
in Figure~\ref{fig:widths}, which shows the median, upper and lower quartiles, 
and upper and lower deciles of their widths for each dataset. Each chart is 
divided into three parts, one for each confidence level we consider, and each
part contains a boxplot for each nonconformity measure used.

The values reported in Tables~\ref{tab:bostonres}-\ref{tab:bankres} in 
conjunction with the range of possible labels of each dataset show that
the prediction intervals produced by our method are quite tight. For a 
confidence level as high as $99\%$ the median widths obtained with 
nonconformity measure (\ref{eq:nm1}) are $87.4\%$, $53.5\%$, $20.1\%$ 
and $28.5\%$ of the label range of the four datasets respectively, while 
the best widths obtained with nonconformity measure (\ref{eq:nm2}) are 
$71.5\%$, $44.5\%$, $17.5\%$ and $16.3\%$ of the label range. 
If we now consider the slightly lower $95\%$ confidence level the 
median widths obtained with nonconformity measure (\ref{eq:nm1}) are 
$38.2\%$, $32.4\%$, $12.5\%$ and $17.5\%$ of the label ranges respectively, 
while the best widths obtained with nonconformity measure (\ref{eq:nm2}) 
are $35.8\%$, $26.1\%$, $11.1\%$ and $10\%$ of the label ranges.
It is worth to note that the relatively
big intervals obtained for the Boston Housing dataset at the $99\%$ 
confidence level are partly due to the small size of the calibration 
set used in this case; this is also the reason for which these 
intervals become much tighter at the $95\%$ confidence level. 
Figure~\ref{fig:widths} shows the difference between the two 
nonconformity measures and demonstrates the improvement achieved by 
our normalized nonconformity measure (\ref{eq:nm2}). With the exception 
of the Boston Housing 
dataset, in all other cases more than half of the intervals of 
measure (\ref{eq:nm2}) are tighter than the 25th percentile of the 
widths produced by measure (\ref{eq:nm1}). In the case of the Bank 
dataset the difference is even more impressive. The same figure also
shows the difference between the two values of $\beta$. When $\beta$ 
is set to $0$ the width distribution is bigger as the measure
is more sensitive. When we slightly increase $\beta$ the sizes of 
the widths fluctuate less and in most cases become generally a bit 
smaller. In the case of the Bank dataset the value of $\beta = 0.5$ 
was clearly too large considering that the label range of the dataset 
was smaller than $0.5$.

Finally, the values reported in the rightmost part 
of Tables~\ref{tab:bostonres}-\ref{tab:bankres} demonstrate the reliability 
of the obtained prediction intervals. The percentages reported in this 
part of the four tables are in all cases very near the required 
significance levels.

\section{Total Electron Content Prediction}\label{sec:tec}

\subsection{Characteristics and Measurement Data}

TEC is defined as the total amount of electrons along a particular line of
sight in the ionosphere and is measured in total electron content units (1 TECu = $10^{16}$ el/m$^2$).
It is an important parameter in trans-ionospheric links since when
multiplied by a factor which is a function of the signal frequency,
it yields an estimate of the delay imposed on the signal by the ionosphere
(an ionized region ranging in height above the surface of the earth
from approximately 50 to 1000 km) 
due to its dispersive nature~\citep{kersley:tec}.
The necessity for accurate prediction of TEC stems out of the fact that 
up to date and accurate information is needed in the application of mitigation 
techniques for the reduction of ionospheric imposed errors on radar, 
communication, surveillance and navigation systems.

\begin{figure}[t]
	\centering
		\subfloat[Noon values of TEC]{\includegraphics[trim=0mm 4mm 0mm 4mm,clip,width=8cm]{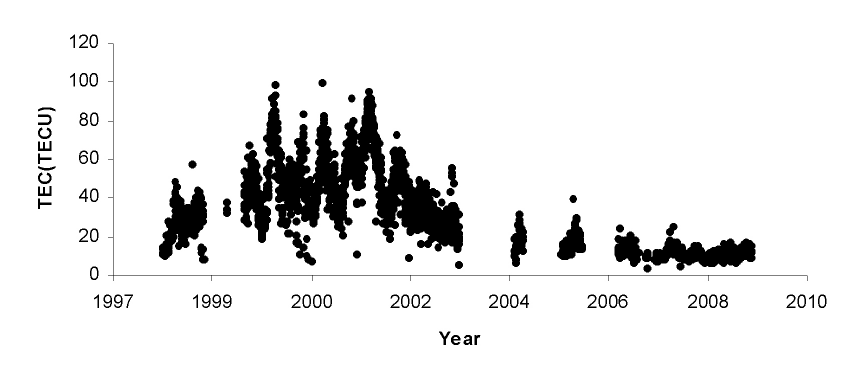}\label{fig:longtermvar}} \\
		\subfloat[Modelled monthly mean sunspot number]{\includegraphics[trim=0mm 4mm 0mm 4mm,clip,width=8cm]{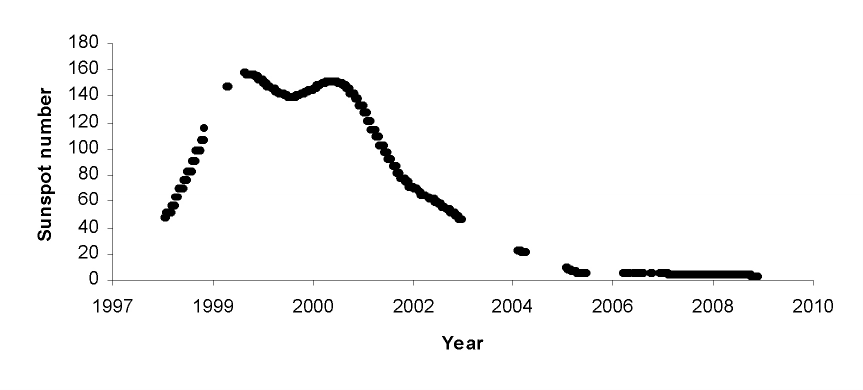}\label{fig:sunspot}}
\caption{Long-term variability of TEC and solar activity.}
\label{fig:LongTermVar}
\end{figure}

\begin{figure}
	\centering
		\subfloat[24-hour variability]{\includegraphics[width=5.5cm]{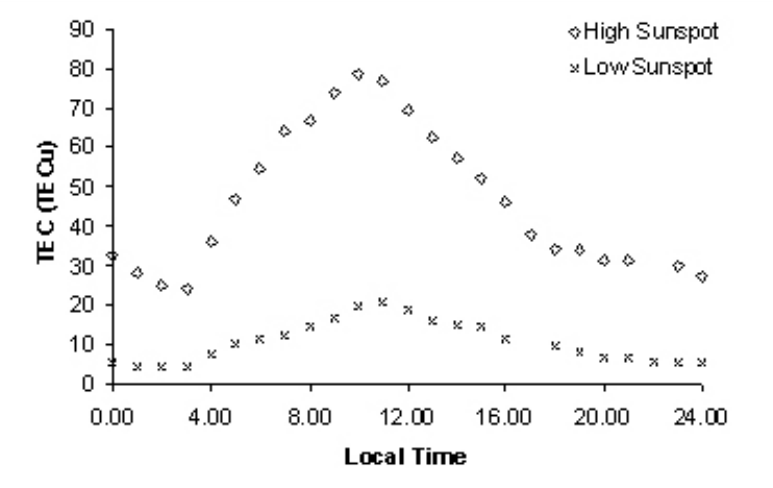}\label{fig:24hourVar}}
		\subfloat[Seasonal variability]{\includegraphics[trim = 5mm 0mm 7mm 0mm, clip, width=8cm]{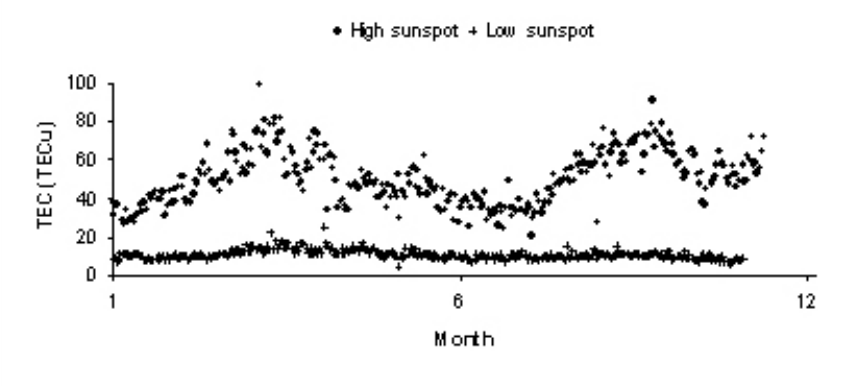}\label{fig:SeasonalVar}}
\caption{24-hour and seasonal variability of TEC for low and high solar activity.}
\label{fig:24hourSeasonalVar}
\end{figure}

The density of free electrons within the ionosphere and 
therefore TEC depend upon the strength of the solar ionizing radiation 
which is a function of time of day, season, geographical location and 
solar activity~\citep{goodman:hf}. The long-term effect of solar 
activity on TEC follows an eleven-year cycle and is shown in 
Figure~\ref{fig:LongTermVar} where noon values of TEC are plotted against time 
as well as a modelled monthly mean sunspot number (R), which is a well 
established index of solar activity. Comparing the two we can observe 
a clear correlation between the mean level of TEC and the modelled sunspot number.
We should note that during the last couple of years solar activity was characterised 
by a prolonged period of low sunspot values (see Figure~\ref{fig:LongTermVar}). This 
however does not pose a problem to the approach adopted in this paper as the 
cyclic variation of solar activity is not embedded in the model specification. 
The 24-hour variability of TEC is demonstrated with two examples recorded during 
low and high solar activity in Figure~\ref{fig:24hourVar}. Examples 
of its seasonal variability are shown in Figure~\ref{fig:SeasonalVar}, in which 
the noon values of TEC are plotted again for low and high sunspot. As can be 
observed from these figures solar activity has an important effect on both
the 24-hour and seasonal variability of TEC.

The TEC measurements used in this work consist of a bit more than 60000
values recorded between 1998 and 2009.
The parameters used as inputs for modelling TEC are the hour, 
day and monthly mean sunspot number. The first two were converted into
their quadrature components in order to avoid their unrealistic discontinuity
at the midnight and change of year boundaries. Therefore the following
four values were used in their place:
\begin{equation}\label{eq:sinhour}
	sinhour = {{{{{{{{{\rm sin}}}}}}}}}(2\pi \frac{hour}{24}),
\end{equation}
\begin{equation}\label{eq:coshour}
	coshour = {{{{{{{{{\rm cos}}}}}}}}}(2\pi \frac{hour}{24}),
\end{equation}
\begin{equation}\label{eq:sinday}
	sinday = {{{{{{{{{\rm sin}}}}}}}}}(2\pi \frac{day}{365}),
\end{equation}
\begin{equation}\label{eq:cosday}
	cosday = {{{{{{{{{\rm cos}}}}}}}}}(2\pi \frac{day}{365}).
\end{equation}
It is worth to note that in ionospheric work solar activity is usually 
represented by the 12-month smoothed sunspot number, which however 
has the disadvantage that the most recent value available corresponds 
to TEC measurements made six months ago. In our case in order to enable 
TEC data to be modelled as soon as they are measured, and for future 
predictions of TEC to be made, the monthly mean sunspot number values 
were modelled using a smooth curve defined by a summation of sinusoids.

\subsection{Experiments and Results}\label{sec:exp2}

Our experiments followed the same experimental setting described in
Section~\ref{sec:exp}. Before conducting our experiments all 
attributes of the dataset were normalized to a minimum 
value of $-1$ and a maximum value of $1$. A 2-fold cross-validation 
process was performed 10 times on random permutations of the dataset 
with the 2 folds consisting of 30211 and 30210 examples 
respectively. This allowed the evaluation of the proposed method on the 
whole range of possible sunspot values (which typically exhibit an 11 year cycle), 
since solar activity has a strong effect on the variability of TEC. 
The calibration set size was set to $999$ examples, which resulted in 
$q+1$ in (\ref{eq:ICPregions-s}) being $1000$.

The underlying NN had the same structure and followed the same training 
process as the one used in the experiments of Section~\ref{sec:exp}. In 
this case the number of hidden neurons was set to 13, which was as before 
determined by trial and error. Finally, for nonconformity measure (\ref{eq:nm2}) 
we again experimented with $\beta = 0$ and $\beta = 0.5$.

\begin{table}\footnotesize
  \centering
  \begin{tabular}{|c|rrr|rrr|rrr|} \hline
             & \multicolumn{3}{|c|}{} 
             & \multicolumn{3}{|c|}{Interdecile} & \multicolumn{3}{|c|}{Percentage} \\
             \multicolumn{1}{|c|}{Nonconformity} & \multicolumn{3}{|c|}{Median Width} 
             & \multicolumn{3}{|c|}{Mean Width} & \multicolumn{3}{|c|}{of Errors (\%)} \\
             \multicolumn{1}{|c|}{Measure} & \multicolumn{1}{|c}{90\%} & \multicolumn{1}{c}{95\%}
             & \multicolumn{1}{c|}{99\%} 
             & \multicolumn{1}{|c}{90\%} & \multicolumn{1}{c}{95\%}
             & \multicolumn{1}{c|}{99\%}
             & \multicolumn{1}{|c}{90\%} & \multicolumn{1}{c}{95\%}
             & \multicolumn{1}{c|}{99\%} \\ \hline
             (\ref{eq:nm1})                 & 16.15 & 21.88 & 38.17 & 16.32 & 22.02 & 38.67 & 10.12 & 5.02 & 1.01 \\
             (\ref{eq:nm2}) - $\beta = 0$   & 13.04 & 16.24 & 25.91 & 14.20 & 17.69 & 28.27 & 9.73  & 4.92 & 1.00 \\
             (\ref{eq:nm2}) - $\beta = 0.5$ & 12.90 & 16.17 & 26.81 & 13.82 & 17.33 & 28.70 & 9.82  & 4.96 & 0.97 \\ \hline  
  \end{tabular}
  \caption{Tightness and empirical reliability results.}
  \label{tab:res}
\end{table}
\normalsize

\begin{figure}
  \centering
    \includegraphics[trim = 2mm 6mm 0mm 0mm, clip, width=6.5cm]{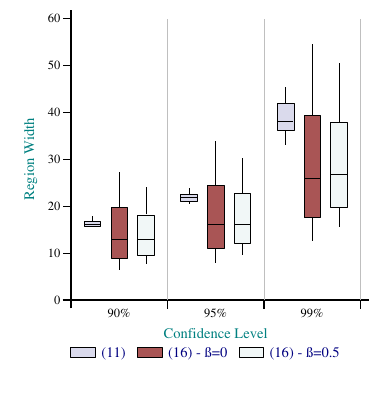}
  \caption{Prediction interval width distribution.}
  \label{fig:tecres}
\end{figure}

\begin{figure}[t]
	\centering
		\subfloat[Low Sunspot]{\includegraphics[trim = 0mm 6mm 0mm 4mm, clip, width=10cm]{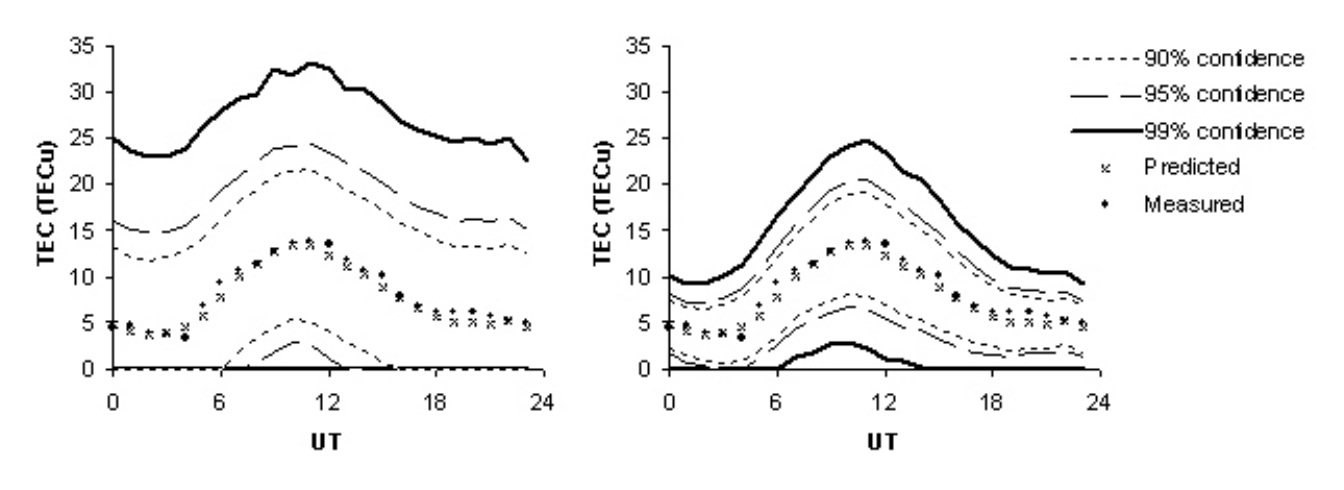}\label{fig:reslow}}\\
		\subfloat[Medium Sunspot]{\includegraphics[trim = 0mm 6mm 0mm 4mm, clip, width=10cm]{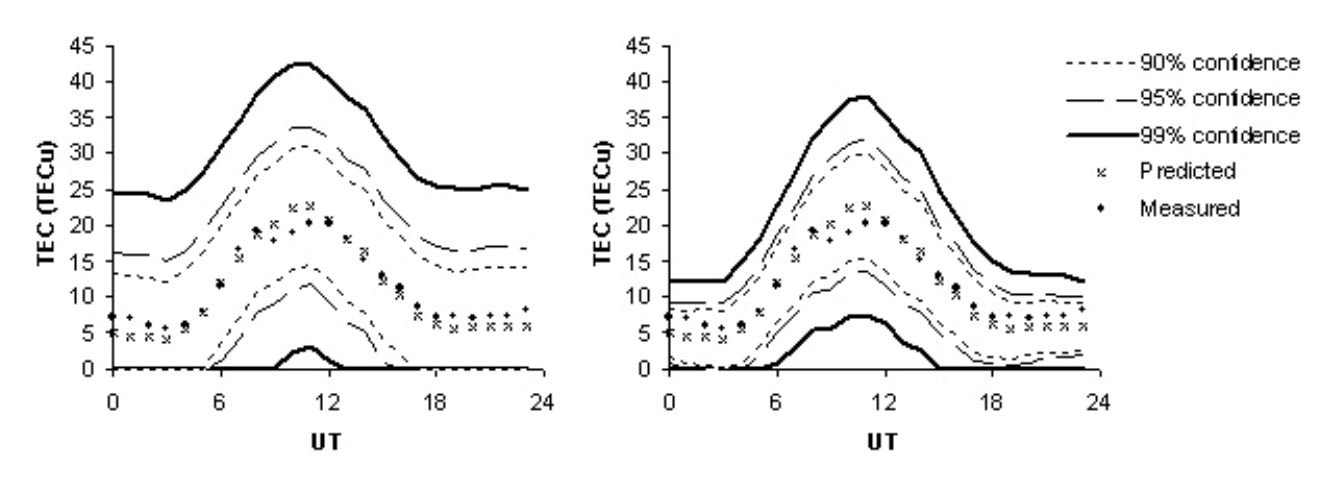}\label{fig:resmed}}\\
		\subfloat[High Sunspot]{\includegraphics[trim = 0mm 6mm 0mm 4mm, clip, width=10cm]{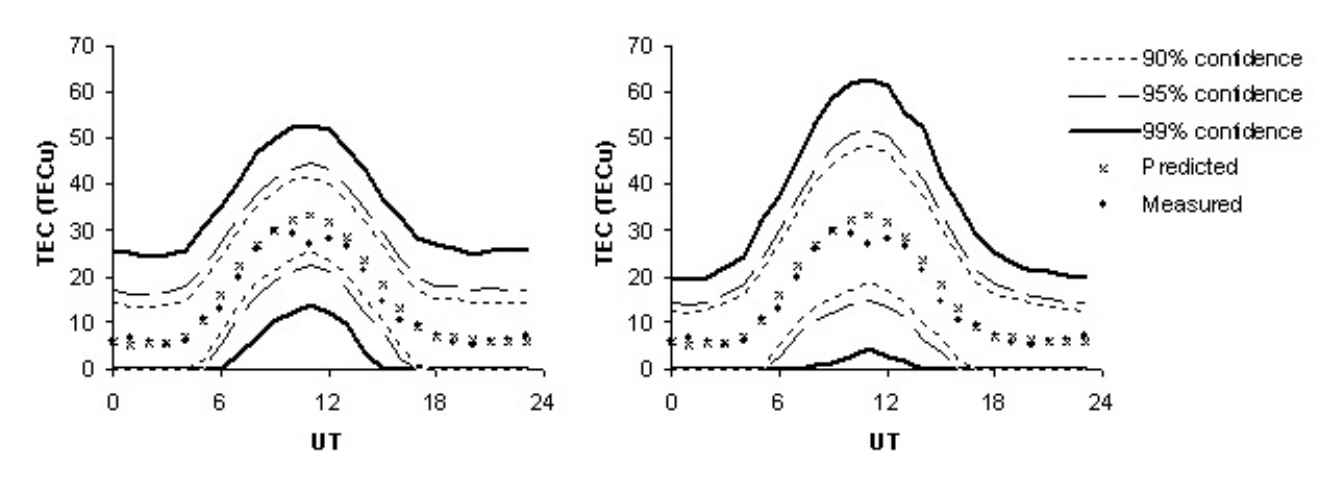}\label{fig:reshigh}}\\
\caption{Examples of the prediction intervals produced by nonconformity measure (\ref{eq:nm1}) 
         on the left and (\ref{eq:nm2}) on the right for typical days in low, medium and high sunspot periods.}
\label{fig:tecexamples}
\end{figure}

In terms of point predictions both our method and the original NN
performed quite well with a RMSE of $5.5$ TECu and a correlation coefficient
between the predicted and the actual values of $0.94$; there was almost no
difference between the two due to the large size of the dataset. The results 
of our method in terms of prediction interval tightness and empirical 
reliability are presented in Table~\ref{tab:res}, while boxplots showing the 
distribution of the obtained prediction interval widths are displayed in
Figure~\ref{fig:tecres}. By considering the range of the measured values 
in our dataset, which are between $0$ and $110$ TECu, we can see that 
the prediction interval widths produced by the proposed method are quite 
impressive. For example, the median width obtained with nonconformity 
measure (\ref{eq:nm2}) and $\beta = 0$ for the $99\%$  confidence level
covers only $23.5\%$ of this range, while for the $95\%$ confidence level
it covers $14.8\%$. It is worth to mention that, since the produced 
intervals are generated based on the size of the absolute error that the 
underlying algorithm can have on each example, a few of the intervals 
start from values below zero, which are impossible for the particular 
application. So we could in fact make these intervals start at zero without
making any additional errors and this would result in slightly smaller values 
than those reported in this table. We chose not to do so here in order to
evaluate the actual intervals as output by our method without any intervention.
The error percentages reported in Table~\ref{tab:res} again demonstrate 
the reliability of the obtained intervals as they are almost equal to the 
required significance level in all cases.

By comparing the boxplots presented in Figure~\ref{fig:tecres} for each of 
the two nonconformity measures we can see the remarkable improvement that 
the normalized nonconformity measure~(\ref{eq:nm2}) achieves. The majority
of the prediction interval widths produced by measure~(\ref{eq:nm2}) are 
below the 10th percentile of those produced by measure~(\ref{eq:nm1}).
The difference between the two measures is further demonstrated in Figure~\ref{fig:tecexamples},
which plots the intervals obtained by each measure for three typical days 
in the low, medium and high sunspot periods. Here we can see that unlike the 
intervals of measure (\ref{eq:nm1}), those of measure (\ref{eq:nm2}) are 
wider at noon and in the high sunspot period, when the variability of TEC 
is higher, but they are much smaller during the night and in the low 
sunspot period.

\section{Conclusions and Future Work}\label{sec:conc}

We have developed a regression ICP based on NNs, which is one of the 
most popular techniques for regression problems. Unlike conventional 
regression NNs, and in general machine learning methods, our algorithm
produces prediction intervals that satisfy a required confidence level.
Our experimental results on four benchmark datasets and on the problem 
of TEC prediction show that the 
prediction intervals produced by the proposed method are not only
well-calibrated, and therefore highly reliable, but they are also 
tight enough to be useful in practice. Furthermore, we defined a 
normalized nonconformity measure, which achieved an impressive 
improvement in terms of prediction interval tightness over the typical 
regression measure.

Our main direction for future research is the development of more 
normalized measures, which will hopefully give even
tighter intervals. Moreover, our future plans include 
the application and evaluation of the proposed method on other 
problems for which provision of prediction intervals is highly
desirable. We also intend to apply Conformal Prediction for investigating 
TEC variability under increased geomagnetic activity, an issue that was
not considered in this paper.

\end{document}